\documentclass[runningheads]{llncs}

\usepackage[mobile]{eccv}

\usepackage{eccvabbrv}

\usepackage{graphicx}
\usepackage{booktabs}

\usepackage[accsupp]{axessibility}  

\usepackage{hyperref}

\usepackage{amssymb}
\usepackage{tikz}
\usepackage{booktabs} 
\usepackage{multirow}
\usepackage{tabularx}  
\usepackage{adjustbox}
\usepackage[ruled,vlined]{algorithm2e}
\usepackage{caption}
\usepackage{colortbl}
\usepackage{setspace}

\definecolor{denim}{rgb}{0.08, 0.38, 0.74}
\definecolor{ava}{RGB}{150, 0, 255}
\definecolor{coralred}{rgb}{1.0, 0.25, 0.25}
\definecolor{aquamarine}{rgb}{0.5, 1.0, 0.83}
\definecolor{green(munsell)}{rgb}{0.0, 0.66, 0.47}
\definecolor{cadmiumgreen}{rgb}{0.0, 0.42, 0.24}
\definecolor{britishracinggreen}{rgb}{0.0, 0.26, 0.15}
\definecolor{debianred}{rgb}{0.84, 0.04, 0.33}

\newcommand{\tp}{{TreeProbe}}

\newcommand{\green}[1]{\textcolor{britishracinggreen}{#1}}
\usepackage{dsfont}
\usepackage{wrapfig}

\usepackage{enumitem}
\setlist[itemize,1]{label=\textbullet}

\usepackage[dvipsnames]{xcolor}
\newcommand{\red}[1]{{\color{red}#1}}

\newcommand{\weighting}{OCW}

\usepackage{orcidlink}

\begin{document}

\title{Anytime Continual Learning for Open Vocabulary Classification} 


\author{Zhen Zhu\orcidlink{0000-0003-1557-8473} \and Yiming Gong\orcidlink{0009-0007-5520-7999} \and Derek Hoiem\orcidlink{0000-0001-6260-5708}}



\institute{University of Illinois at Urbana-Champaign\\
\email{\{zhenzhu4,yimingg8,dhoiem\}@illinois.edu}}

\maketitle

\begin{abstract}
    We propose an approach for anytime continual learning (AnytimeCL) for open vocabulary image classification. The AnytimeCL problem aims to break away from batch training and rigid models by requiring that a system can predict any set of labels at any time and efficiently update and improve when receiving one or more training samples at any time. Despite the challenging goal, we achieve substantial improvements over recent methods. We propose a dynamic weighting between predictions of a partially fine-tuned model and a fixed open vocabulary model that enables continual improvement when training samples are available for a subset of a task's labels. We also propose an attention-weighted PCA compression of training features that reduces storage and computation with little impact to model accuracy. Our methods are validated with experiments that test flexibility of learning and inference. Code is available at \href{https://github.com/jessemelpolio/AnytimeCL}{https://github.com/jessemelpolio/AnytimeCL}.
    \keywords{Continual learning \and Anytime learning \and Open-vocabulary classification}
\end{abstract}

\section{Introduction}

Continual learning aims to improve a system’s capability as it incrementally receives new data and labels, which gains increasing importance with the expanding scale and applications of visual learning. Continual learning for classification is traditionally approached with discrete label spaces. Adding labels or tasks over time changes the problem landscape, thereby increasing the difficulty of learning.
The open vocabulary setting, however, frames classification as comparing continuous feature and label embeddings. Any images and textual labels can be embedded, so learning in this setting involves only improving on the existing problem definition, which may be more amenable to continuous improvement. Although open vocabulary models can predict over arbitrary label sets, even models like CLIP~\cite{radford_icml2021_clip} that are trained on internet-scale data have unsatisfactory performance on many tasks~\cite{Tong_NeurIPS_2023_MLM_failures,Tong_2024_eyes_wide_shut}. 

Our work aims to continually improve open vocabulary image classifiers as new labeled data is received. We call this ``{\bf anytime continual learning}'' because the goal is to improve efficiently {\em any time} new examples are received and to maintain the ability to predict over arbitrary label sets at {\em any time}. 

Recent work by Zhu~\etal.~\cite{treeprobe} extends CLIP by training linear classifiers and combining their predictions with those of label vectors from the original text embedding. To enable efficient and distributed learning, linear classifiers are trained for each partition of the feature space defined by online hierarchical clustering. This approach requires storing all training examples, but a new example is efficiently incorporated by training only on the data from the example's nearest cluster. In experiments on their own benchmark designed for open vocabulary continual learning and existing benchmarks, Zhu~\etal.~\cite{treeprobe} outperform other recent approaches.

\begin{wrapfigure}{r}{0.5\textwidth}
 \centering
\includegraphics[width=\linewidth]{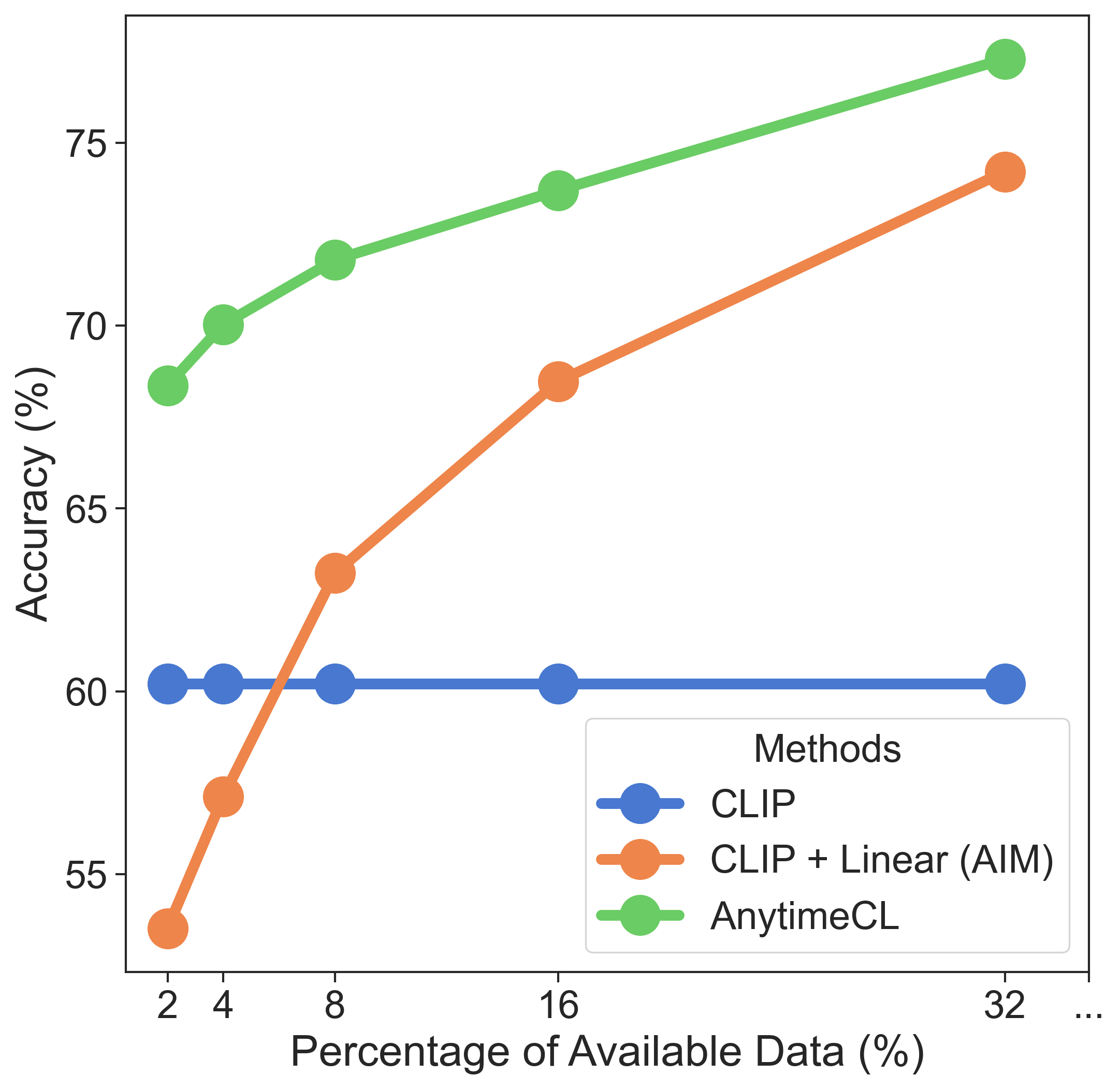}
\caption{
\label{fig:anytime_cl} Our AnytimeCL algorithm can be efficiently updated with each new example and continuously improve. By dynamically weighting predictions between a tunable model and a frozen open vocabulary model, our method can predict over any label set while gaining expertise. 
This figure shows our method outperforms previous SotA Zhu~\etal.~\cite{treeprobe} in every stage in the data incremental setting. 
Our method also outperforms in other settings like task-incremental and class-incremental.
}
\end{wrapfigure}

Our approach is to train online, fine-tuning the last transformer block while keeping the label embedding fixed. When a new training sample is received, we fill a class-balanced batch with stored samples and update the last transformer block in a single step. Our experiments show this performs better than retraining a classifier layer, and each new sample can be incorporated in milliseconds. 
Additionally, we introduce a modified loss, enabling the prediction of ``none of the above'' when the true label is absent from the candidate set, which improves overall performance.

Whenever a new training example arrives, our method updates an estimate of the tuned and original model's accuracy for the given label. The tuned and original model's predictions are then weighted in proportion to their expected accuracy for each label. This weighting accounts for the growing effectiveness of the tuned model, and greatly outperforms Zhu~\etal.~\cite{treeprobe}'s AIM weighting based on whether an example's label is likely to have been seen in training.


The partial fine-tuning approach requires storing either full images that need to be re-encoded during training, or data-consumptive patch-level features at intermediate layers. We apply a per-image weighted PCA, which provides a 30x reduction of data with little impact on prediction accuracy. Such compression could be especially beneficial in a large-scale federated learning setting.


In experiments on the open vocabulary continual learning benchmark proposed by Zhu~\etal.~\cite{treeprobe}, our approach outperforms under all settings and stages, including data-incremental, class-incremental, and task-incremental learning, and zero shot prediction.  Our ablations evaluate the effects of all key design parameters, including the partial fine-tuning, model prediction weighting, batch sampling, and the regularization loss term. We also show that our method can be applied to combine non-open vocabulary models like DINOv2 with CLIP to improve further while maintaining open vocabulary ability.

In summary, our \textbf{main contribution} is an open vocabulary continual learning method that can quickly incorporate new training examples received in any order and continually improve while maintaining open vocabulary performance. Our experiments demonstrate that our system's accuracy and efficiency stem from multiple proposed innovations:

\begin{itemize}
    \item Partial fine-tuning of features with a fixed label embedding;
    \item Online training with each batch composed of the new training sample and class-balanced stored samples;
    \item Online learning of per-label accuracy for effective combination of original and tuned model predictions;
    \item Loss modification to enable ``none of the above'' prediction, which also stabilizes open vocabulary training;
    \item Intermediate layer feature compression that reduces storage of training samples and improves speed without much loss to accuracy.
\end{itemize}


\section{Related work}



{The goal of continual learning (CL) is to continually improve the performance when seeing more data, while online continual learning~\cite{GEM,Aljundi_2019_GSS,Aljundi_2019_MIR,prabhu_gdumb_eccv_2020,Caccia_ICLR_2022_AML,Prabhu_arxiv2023_online_continual} aims to achieve a good trade-off between performance and learning efficiency. } Our problem setup and goals are distinguished by these important characteristics:
\begin{itemize}
\item {\em Flexibly learn}, receiving labeled examples in data-incremental (random order), class incremental (grouped by category), or task incremental (grouped by dataset);
\item {\em Efficiently update} from a single example or a batch of examples;
\item {\em Retain training data}, though potentially in compact, privacy-preserving forms;
\item {\em Cumulatively improve} from new data without forgetting other classes or tasks; 
\item {\em Flexibly infer}, predicting over label sets defined at inference time, without rigid task boundaries. 
\end{itemize}

Continual learning approaches can be broadly categorized into regularization~\cite{LwF,EWC,SI}, replay/rehearsal methods~\cite{iCaRL,DER,GEM,generativereplay}, and parameter isolation or expansion~\cite{ExpertGate,Rusu16Progressive,hardattention,Mallya18Packnet,sidetuning,yoon_lifelong_dynamically_expandable,prabhu_gdumb_eccv_2020,prabhu_budgetedCL_cvpr_2023}. Regularization techniques generally impose constraints on the learning process to alleviate forgetting. Rehearsal methods involve storing and replaying past data samples during training~\cite{generativereplay,rainbowmemory,prabhu_gdumb_eccv_2020,prabhu_budgetedCL_cvpr_2023}. So far, when training data can be stored, simple replay methods tend to outperform others, whether limiting the number of stored examples~\cite{prabhu_gdumb_eccv_2020} or training computation~\cite{prabhu_budgetedCL_cvpr_2023}.
Parameter isolation methods maintain learning stability by fixing subsets of parameters~\cite{Mallya18Packnet,hardattention} or extending model with new parameters~\cite{Rusu16Progressive,yoon_lifelong_dynamically_expandable,sidetuning}. Recently, several works~\cite{wang_2022_CVPR_l2p,dualprompt_wang_eccv2022,Jin_2022_ACL_good_prompt,Wang_2022_nips_sprompt,Khattak2023multimodalprompt,Khattak2023selfregulatingprompts,coda_prompt_Smith_2023_CVPR} adopt prompt tuning for continual learning, which can also fall into the parameter expansion category. These methods can be used for stage-wise incremental continual learning, but none demonstrates the efficacy on online incremental learning that is our focus. Prompt tuning provides an alternative to weight-based fine-tuning with less forgetting. We consider weight-tuning approaches here, but preliminary experiments indicate the same strategy is applicable to prompt tuning, with similar accuracy but slower training.

We now focus on those most relevant and influential to our work; see Wang et al.~\cite{wang_2023_comprehensive_continual_learning_survey} for a comprehensive survey of continual learning.


\subsection{CL for open-vocabulary classification}

WiSE-FT~\cite{WiSE-FT} fine-tunes CLIP encoders on target tasks and averages fine-tuned weights with original weights for robustness to distribution shifts. PAINT~\cite{Ilharco2022patching} shares a similar idea as WiSE-FT but the weight mixing coefficient is found via a held-out set. With such approaches, generality degrades with each increment of continual learning. CLS-ER~\cite{CLS_ER} exponentially averages model weights in different paces for its plastic and stable model to balance learning and forgetting. 
Closer to our work, ZSCL~\cite{ZSCL} fine-tunes CLIP encoders using a weight ensemble idea, similar to WiSE-FT, and applies a distillation loss with a large unsupervised dataset to reduce forgetting. Unlike these, we partially fine-tune CLIP with fixed label embeddings and use a dynamically weighted combination of the predictions from the tuned and original models, which enables improvement on target tasks without sacrificing the generality of the original model.

Another class of methods~\cite{skorokhodov_2021_generalized_zero_shot_iclr_2021,Gautam_2021_CZSL_Online_Generalized_zero_shot} use attribute-based recognition~\cite{farhadi2009describing,lampert_attribute_2014_pami}, continually learning attributes and attribute-class relations as a way to generalize from seen to unseen categories. We use CLIP, which provides zero-shot ability by training to match image-text pairs on a large training corpus, instead of the attribute-based approach.  In reported results~\cite{radford_icml2021_clip,skorokhodov_2021_generalized_zero_shot_iclr_2021,Gautam_2021_CZSL_Online_Generalized_zero_shot} and our own tests, CLIP has much higher zero-shot accuracy (e.g. for the SUN and aPY datasets) than the continual attribute-based methods achieve even after training, supporting the idea of building on vision-language models for open vocabulary image classification.

\subsection{CL with constrained computation}
Prabhu \etal.~\cite{prabhu_budgetedCL_cvpr_2023} evaluate a wide variety of continual learning techniques under a setting of retaining data but limiting compute. Compared to all other techniques, they find that simple replay strategies, such as fine-tuning with uniformly random or class-balanced batches, are most effective in this setting. They also provide some evidence that partial-fine tuning of a well-initialized network may be an efficient and effective approach. In this direction, we find that fine-tuning only the last transformer block of CLIP by performing one mini-batch update per incoming sample works surprisingly well.


\subsection{Biologically inspired CL}

The complimentary learning systems theory~\cite{cls_oreilley_2014} (CLS theory) posits that humans achieve continual learning through the interplay of sparse retrieval-based and dense consolidated memory systems.  This has inspired many works in continual learning, e.g.~\cite{pham_dualnet_fast_slow_2021,CLS_ER,treeprobe}. The most closely related is Zhu \etal.~\cite{treeprobe}, as detailed in the introduction.  Many works are also inspired by the idea of wake/sleep phases of learning, where faster updates are required during the wake stage.  
Recent studies of mammalian memory and learning indicate that consolidation occurs in similar ways via replay during wakefulness and sleep~\cite{consolidation_siegel_2021,consolidation_wamsley_2022,consolidation_brodt_2023}. {Our online learning method mixes use of individual examples and consolidated networks for continuous improvement for continuous improvement, reflecting the idea of wakeful consolidation through replay.}

\subsection{Memory compression}

Other work has compressed training data~\cite{AQM_icml_2020}, or distilled training data into a network~\cite{yu2023_dataset_distillation_review}, or learned prompts~\cite{wang_2022_CVPR_l2p,dualprompt_wang_eccv2022,Jin_2022_ACL_good_prompt,Wang_2022_nips_sprompt,Khattak2023multimodalprompt,Khattak2023selfregulatingprompts,coda_prompt_Smith_2023_CVPR} instead of maintaining original examples. AQM~\cite{AQM_icml_2020} uses an online VQ-VAE~\cite{vqvae_neurips2027_van_oord} to compress training samples to reduce storage for continual learning.  We instead compress intermediate features to improve both storage and computation of partial fine-tuning. 

\section{Method}

\begin{figure*}[!htb]
 \centering
 \begin{minipage}{\textwidth}{
\includegraphics[width=\linewidth]{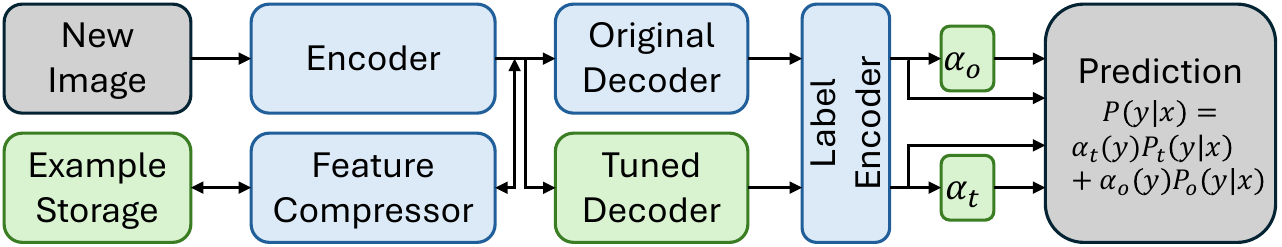}
\caption{Overview. On receiving a new training sample, the batch is completed with stored samples, the prediction is made using a label embedding, the tuned decoder and confidence weights ($\alpha_o$, $\alpha_t$) are updated in one step, and the new sample is stored. To save space and time, stored examples are encoded and compressed. In testing, the probability of each candidate label is determined by predictions from both decoders and their class-wise confidence weights. Our method enables constant-time updates from new samples while continually improving and maintaining open vocabulary performance. Green blocks are updated during training; blue blocks are not updated.}
\label{fig:overview}
}
\end{minipage}
\end{figure*}

Our system receives training examples ($x$, $y$, $\mathcal{Y}$) one by one to update its models, in order to continually improve in prediction of $y$ given ($x$, $\mathcal{Y}$), where $x$ is an input image, $y$ is its target label, and $\mathcal{Y}$ is its set of candidate labels.  


Fig.~\ref{fig:overview} offers a system overview aimed at enhancing open vocabulary image classifiers by integrating a tuned model for learning target tasks with a frozen original model. The tuned model uses the same encoder as the original but incorporates a trainable decoder.
For an image $x$, both the tuned model and the original model produce the probabilities for all candidate labels, denoted as $P_t(y|x)$ and $P_o(y|x)$. The final probability for the image is weighted through our Online Class-wise Weighting (OCW, Sec.~\ref{sec:method_weighting}): 
\begin{equation}
    \label{eq:our_weighting}
    P(y|x) = \alpha_o(y) P_t(y|x) + \alpha_t(y) P_o(y|x),
\end{equation} where $\alpha_o(y)$ and $\alpha_t(y)$ are the relative expected accuracies for the original and tuned models for label $y$.  
During training, new samples are encoded to intermediate features (feature vectors for patches plus a CLS token), optionally compressed (Sec.~\ref{sec:method_compression}), and stored, for reuse in future steps.

\subsection{Models}
\label{sec:method_models}

\noindent \textbf{Original model.}~
In our experiments, the original model is the publicly available CLIP~\cite{radford_icml2021_clip} ViT model that has been trained contrastively on image-text pairs. In our experiments, the shared encoder is all but the last transformer block, and the original decoder is the last block. The CLIP model produces a probability for label $y$ for image $x$ given a set of candidate text labels $\mathcal{Y}$ based on the dot product of image embedding $e_{x}$ (CLS token) and text embedding $e_{y}$:  
\begin{equation}
\label{eq:class_wise_probability}
    P_o(y|x) = \frac{\exp(100 \cdot \cos(e_{x}, e_{y}))}{\sum_{y_k\in\mathcal{Y}} \exp(100 \cdot \cos(e_{x}, e_{y_k}))}.
\end{equation}


\noindent \textbf{Tuned model.}~
With each new sample, our goal is to efficiently update the tuned model to improve accuracy in received labels while maintaining general features that are effective for future learning. We tune only the last image transformer block while keeping the label embedding fixed, which helps the features to stay correlated with the text modality and reduces overfitting to received labels. Note that more blocks can be tuned, which could enable better performance but with increased computation. The tuned decoder is initialized with the weights of the original decoder.  

Given a new sample, we form a batch of that sample and a class-balanced sampling of stored training samples. Specifically, we first determine the class count for selection: for a batch size of $B$ and all seen labels $\mathcal{Y}_t$, we uniformly select $\min(B-1, |\mathcal{Y}_t|)$ classes from $\mathcal{Y}_t$, where $|\cdot|$ indicates the length of a set. Then we uniformly sample an equal number of instances from each chosen class to form the batch. This class-balanced sampling ensures continual retention and improvement of prediction of training labels. Then, the tuned model is updated with one iteration based on a cross-entropy loss.

Additionally, we develop a regularization loss that helps with performance. The idea is that, if the true label is not in the label candidates, a low score should be predicted for every candidate label.  We implement this with an ``other'' option within the candidate set, but since ``other'' does not have an appearance, we model it with just a learnable bias term. The combined loss for training the tuned model is therefore: 
\begin{equation}
\label{eq:final_loss}
 \mathcal{L}(x, y, \mathcal{Y}) =\mathcal{L}_{\text{ce}}(x,y,\mathcal{Y} \cup \text{other}) + \beta \mathcal{L}_{\text{ce}}(x,\text{other},(\mathcal{Y} \cup \text{other}) \setminus y),
\end{equation}
where $\beta$ is a hyperparameter balancing the two loss terms, set to $0.1$ based on validation experiments.

\subsection{Online Class-wise Weighting}
\label{sec:method_weighting}

We estimate the probability of each candidate label by weighted voting of the tuned and original models (Eq.~\ref{eq:our_weighting}). We would like to assign more weight to the model that is more likely to be correct for a given label. There are two big problems: 1) training samples would provide a highly biased estimate of correctness for the tuned model; 2) maintaining a held out set or performing cross-validation would either reduce available training samples or be impractically slow. 
Our solution is to use each training sample, before update, to update our estimate of the likelihood of correctness for its label based on the tuned and original predictions.

Let $c_t(y)$ and $c_o(y)$ be the estimated accuracy of the tuned and original model for label $y$. We apply an exponential moving average (EMA) updating method to estimate them online, ensuring the estimations are reliable when evaluated at any time, concurring with our anytime continual learning goals. 
Assuming the EMA decay is set to $\eta$ (=0.99 in our experiments), the estimated accuracy of the tuned model at the current step is:
\begin{equation}
    c_t(y) = \eta \hat{c}_t(y) + (1 - \eta) \mathds{1}[y_t(x)=y].
\end{equation}
Here, $\hat{c}_t(y)$ is the estimated accuracy of label $y$ in the previous step; $y_t(x)$ denotes the predicted label of the tuned model for $x$. Since the exponential moving average depends on past values, we compute $c_t(y)$ as the average accuracy for the first $\lfloor \frac{1}{1-\eta} \rfloor$ samples. $c_o(y)$ is updated in the same way.

After getting $c_t(y)$ and $c_o(y)$, the weights of the two models are: 
\begin{equation}
    \label{eq:final_alpha}
    \alpha_t(y)= \frac{c_t(y)}{c_t(y) + c_o(y) + \epsilon}, \qquad \alpha_o(y)= 1 - \alpha_t(y).
\end{equation}
Here, $\epsilon$ is a very small number (1e-8) to prevent division by zero. For labels not seen by the tuned model, we set $\alpha_t(y)=0$, so $\alpha_o(y)=1$.

\subsection{Storage efficiency and privacy}
\label{sec:method_compression}

Partial tuning of our model requires either storing each image or storing the features (or tokens) that feed into the tuned portion of the model. Storing images has disadvantages of lack of privacy and inefficiency in space and computation, due to need to re-encode in training.
Storing features alleviates some of these problems, 
but still uses much memory or storage. Therefore, we investigate how to compress the training features, aiming to increase storage and computational efficiency while maintaining training effectiveness. The compressed features also provide more privacy than storing original images, though we do not investigate how well the images can be recovered from compressed features.

Well-trained networks learn data-efficient representations that are difficult to compress. Indeed, if we try to compress feature vectors with VQ-VAE~\cite{vqvae_neurips2027_van_oord} or PCA (principal component analysis) trained on a dataset, we are not able to achieve any meaningful compression without great loss in training performance. The features within {\em each image}, however, contain many redundancies. We, therefore, compute PCA vectors on the features in each image and store those vectors along with the coefficients of each feature vector. Further, not all tokens are equally important for prediction. Hence, we train a per-image attention-weighted PCA, weighted by attention between each token and the CLS token. Finally, we can compress further by storing min/max floating point values for each vector and for the coefficients and quantizing them to 8-bit or 16-bit unsigned integers. See the supplemental material for details.  By storing only five PCA vectors and their coefficients this way, we can reduce the storage of fifty 768-dim tokens ($7\times 7$ patch tokens + CLS token) from 153K bytes to 5K bytes with less than 1\% difference in prediction accuracy. 
\section{Experiments}

Using the setup from~\cite{treeprobe}, we sequentially receive training samples for a target task in a {\bf data-incremental} (random ordering) or {\bf class-incremental} (class-sorted ordering). Or, in a {\bf task-incremental} setting, we receive all examples for each task sequentially. Within each set, we incorporate new examples {\bf one by one} in an online fashion. After receiving each set, the model is evaluated on the entire task (including classes not yet seen in training) for data- or class-incremental, or on all tasks (including tasks not yet seen in training) for task-incremental. 

Our main experiments use the same setup as Zhu \etal.~\cite{treeprobe}, which are designed to test flexible learning and flexible inference. A {\bf target task} contains a subset of seen labels while a {\bf novel task} contains none of these labels.  Target tasks are CIFAR100~\cite{cifar_100}, SUN397~\cite{sun397}, FGVCAircraft~\cite{fgvc_aircraft}, EuroSAT~\cite{eurosat}, OxfordIIITPets~\cite{oxfordpets}, StanfordCars~\cite{stanfordcars}, Food101~\cite{Food101}, and Flowers102~\cite{flowers102}. Novel tasks are ImageNet~\cite{ImageNet}, UCF101~\cite{UCF101}, and DTD~\cite{dtd}. Training examples are received for target tasks, but not for novel tasks. Under this setup, we have 226,080 training samples and 1,034 classes in total. 

\begin{itemize}
    \item \textbf{Task Incremental Learning} updates one model with each of the eight target tasks sequentially. 
    \item \textbf{Class Incremental Learning} updates a per-task  model with examples from a subset (one-fifth) of classes for a stage. 
    \item \textbf{Data Incremental Learning} updates a per-task model with a random subset of the data in eight stages: 2\%, 4\%, 8\%, 16\%, 32\%, 64\%, 100\%. 
\end{itemize} 
After each stage, accuracy is averaged across all tasks (including tasks/classes not yet seen).

{Additionally, we provide a {\bf union data incremental} scenario where we mix all target tasks together and union all target labels, mainly for hyperparameter selection and ablation experiments.} 
{\bf Flexible inference} is evaluated after the task-incremental learning for prediction over novel tasks and sets of candidate labels $\mathcal{Y}$: zero-shot, union of all target tasks/labels and zero-shot tasks/labels, and a mix of some target and zero-shot tasks/labels. See~\cite{treeprobe} for details. The union and mixed settings are most challenging because they require effective combined use of both the original and tuned models, without any task identifiers or indication whether an example's label is in the tuned model's domain. 


\noindent \textbf{Implementation details}. 
We employ the ViT-B/32 model from CLIP as our network backbone. We use AdamW~\cite{AdamW} as the optimizer with a weight decay of 0.05, and we adhere to CLIP's standard preprocessing without additional data augmentations. For offline training, following~\cite{CLIPItself}, we set the learning rate at 6e-4 and the batch size at 2048, employing a cosine annealing scheduler for the learning rate with a minimum rate of 1e-6. For online training, we use a batch size of 32 and a learning rate of $9.375e=\frac{32 \times 6e-4}{2048}$, based on adjustments from the offline training hyperparameters. More method and training details are in the supplemental.


\subsection{Main result}

\begin{figure*}[!htb]
 \centering
 \begin{minipage}{\textwidth}
\includegraphics[width=\linewidth]
{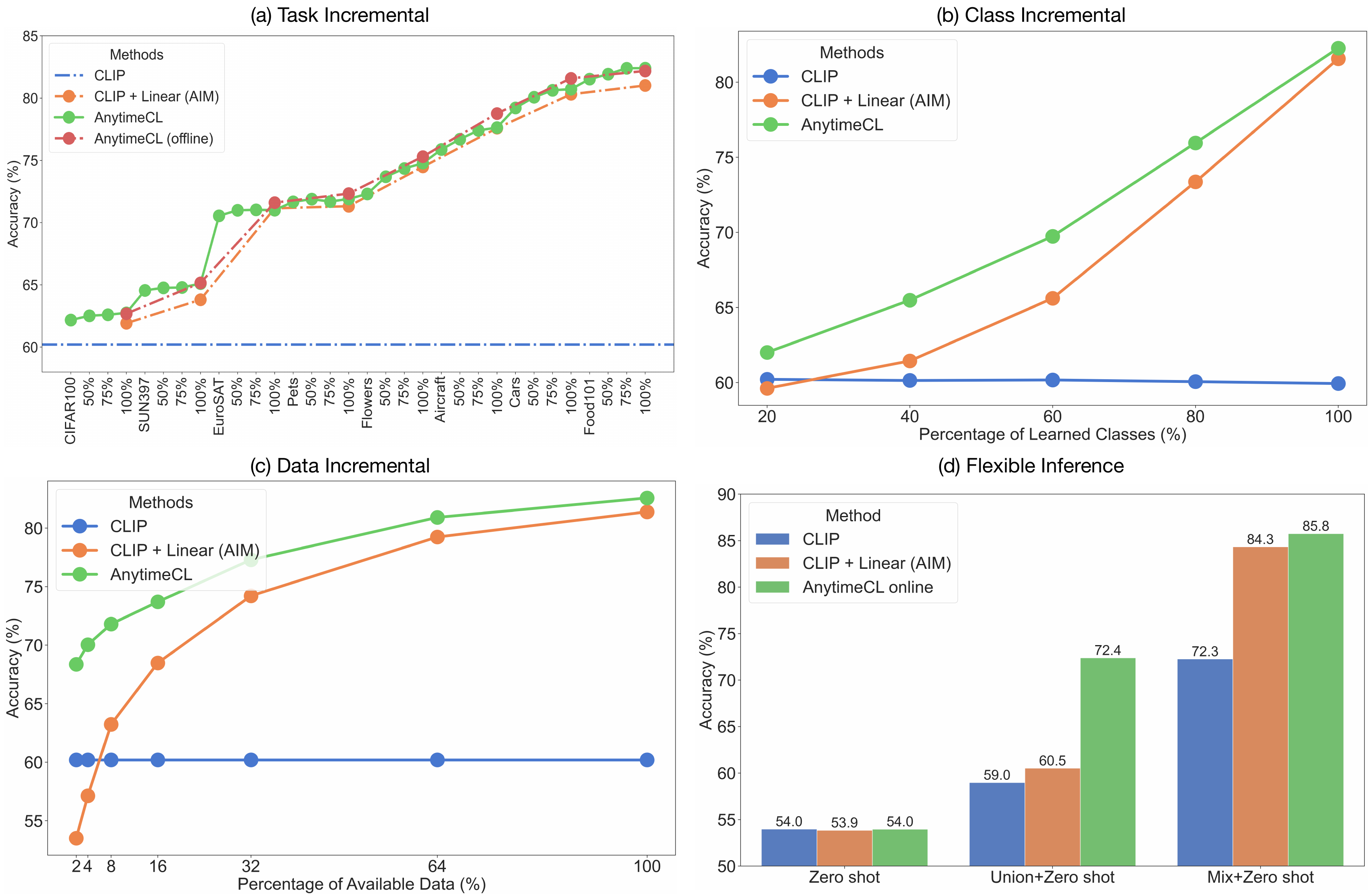}
\caption{Performance comparison of CLIP, CLIP + Linear (AIM), and AnytimeCL variants in task incremental (a); class incremental (b); data incremental (c); and flexible inference (d) settings. In (a), the online approach is represented by a solid line, while offline methods are depicted with dashed lines, assessed when the online algorithms receive 25\%, 50\%, 75\%, and 100\% data of a task, labeling tasks at the 25\% point. 
\label{fig:main_results}
}
\end{minipage}
\end{figure*}

In this evaluation, we compare our method with the previous SotA~Zhu~\etal.~\cite{treeprobe} (CLIP+LinProbe (AIM-Emb)) under the task-, class-, data-incremental scenarios, as well as flexible inference. From all plots in Fig.~\ref{fig:main_results}, our AnytimeCL method consistently outperforms CLIP+LinProbe (AIM). For task incremental, the improvement is mainly due to the partial finetuning with fixed label embeddings. To provide a more direct comparison with Zhu~\etal.~\cite{treeprobe}, we train our approach only after receiving all data of a task (\eg, 100\% CIFAR100) and term this as AnytimeCL (Offline), which also shows steady improvements over Zhu~\etal.~\cite{treeprobe}. For early stages in class incremental and data incremental, the performance improvement mainly comes from the advantage of our online class-wise weighting method over AIM, which only considers whether the testing sample is likely to be from a known class. 

\subsection{Comparison under MTIL task incremental learning~\cite{ZSCL}}
\label{sec:zscl_comparison}
\begin{table*}
\centering
\scalebox{0.95}{
\begin{tabular}{|l|cc|cc|cc|}
\hline
\textbf{Method} & \textbf{Transfer} & $\Delta$ & \textbf{Avg.} & $\Delta$ & \textbf{Last}  & $\Delta$\\
\hline
CLIP & 69.4 & 0.0 & 65.3 & 0.0 & 65.3 & 0.0 \\ \hline
LwF~\cite{LwF} & 56.9 & \red{-12.5} & 64.7 & \red{-0.6} & 74.6 & \green{+9.3} \\
iCaRL~\cite{iCaRL} & 50.4 & \red{-19.0} & 65.7 & \green{+0.4} & 80.1 & \green{+14.8} \\
WiSE-FT~\cite{WiSE-FT} & 52.3 & \red{-17.1} & 60.7 & \red{-4.6} & 77.7 & \green{+12.4} \\
ZSCL~\cite{ZSCL}   & 68.1 & \red{-1.3} & 75.4 & \green{+10.1} & 83.6 & \green{+18.3} \\ \hline
\tp\ (AIM) (50k)~\cite{treeprobe} & {69.3} & {\red{-0.1}} & {75.9} & {\green{+10.6}} & {85.5} & {\green{+20.2}} \\
LinProbe (AIM)~\cite{treeprobe} & {69.3} & {\red{-0.1}} & \textbf{77.1} & \textbf{\green{+11.8}} & {86.0} & {\green{+20.7}} \\
 \hline 
AnytimeCL & \textbf{{69.4}} & \textbf{{\green{0}}} & 77.0 & \green{+11.7} & 85.8 & \green{+20.5} \\
AnytimeCL (Offline) & \textbf{{69.4}} & \textbf{{\green{0}}} & {{77.0}} & {{\green{+11.7}}} & \textbf{{86.2}} & {\green{+20.9}} \\ \hline
\end{tabular}
}
\caption{Comparison of different methods on MTIL in Order I from ZSCL~\cite{ZSCL}. CLIP is our tested zero-shot performance. All other results are taken Zhu~\etal.~\cite{treeprobe} and ZSCL~\cite{ZSCL}'s papers. ``Transfer'' evaluates the model's performance on zero-shot tasks; ``Last'' is the averaged accuracy on all target tasks after finishing the final task while ``Avg.'' computes the average task performance on all training stages.}
\label{tab:anytime_cl_comparison_to_zscl}
\end{table*}

We compare the performance of AnytimeCL with Zhu~\etal.~\cite{treeprobe} and ZSCL~\cite{ZSCL} under the task incremental learning setting as introduced in ZSCL. 
Results are summarized in Tab.~\ref{tab:anytime_cl_comparison_to_zscl}, using the Transfer, Last, and Average metrics from~\cite{ZSCL} for comparison. Improvements over CLIP are denoted in green under the $\Delta$ columns, while declines are marked in red. Our method consistently achieves zero loss in Transfer in both experiments, indicating an absence of forgetting. On this benchmark, our method performs comparably to LinProbe (AIM) and outperforms other methods. This experimental setup does not fully capture our method's key advantages of online training, ability for data and class incremental learning, and open vocabulary prediction. Also, in this setting, a simple weighting method is effective --- to use the tuned model's prediction for tasks with all candidate labels in the training set, or the original model's prediction otherwise.

\subsection{Training features compression}

We test our per-image attention-weighted PCA compresssion for training features (Table~\ref{tab:compression}).  We compare the data size, average full time to process one 32-sample batch (including data loading, uncompressing, and forward/backward training passes), and final accuracy after completing fine-tuning.  This test is on CIFAR100. Compared to processing the full image or full features, using our compressed features saves 30x the storage while achieving nearly the same accuracy. We were not able to achieve good accuracy when compressing with VQ-VAE, even when using large vocabularies, nor when using PCA vectors computed over the entire dataset.  Computing PCA vectors per instance (over the 50 tokens) and weighting the tokens by their CLS-attention each improve the tuned accuracy.
Quantizing the PCA vectors and coefficients further saves space at a small loss to accuracy. Using the compressed features also gives a nearly 2x speedup vs. using the full features (because the compressed features fit entirely in memory) and 6x vs. using the full image, though all times are fast. {We provide more method details in the supplemental along with a comparison of using different per-instance PCA compression methods under the union data incremental scenario, to show how these compression methods influence the result at various timesteps.}

\begin{table}
\centering
\scalebox{0.9}{
\begin{tabular}{|l|c|c|c|}
\hline
\textbf{Compression} & \textbf{KB/example} & \textbf{ms/batch} & \textbf{FT Accuracy} \\
\hline
Full image & 150.5 & 43.9 & 77.8 \\
Full features & 153.6 & 25.6 & 77.8 \\
VQ & 0.2 & 4.5 & 64.8 \\
Dataset-wide PCA (200 components) & 40.2 & 7.6 & 76.4 \\
Per-instance PCA (5 components) & 19.4 & 7.7 & 74.6 \\
+ CLS-weight (5 components) & 19.4 & 8.9 & 77.9 \\
+ int-quantization (5 components) & 5.3 & 13.9 & 77.5 \\
\hline
\end{tabular}
}
\caption{Comparison of different compression methods in terms of memory usage, processing speed, and test accuracy on CIFAR100. See supplementary material for details on how to get ms/batch and FT Accuracy. We use the same seed for all methods.}
\label{tab:compression}
\end{table}

\subsection{Ablations}
\begin{figure*}[!htb]
 \centering
 \begin{minipage}{\textwidth}
\includegraphics[width=\linewidth]{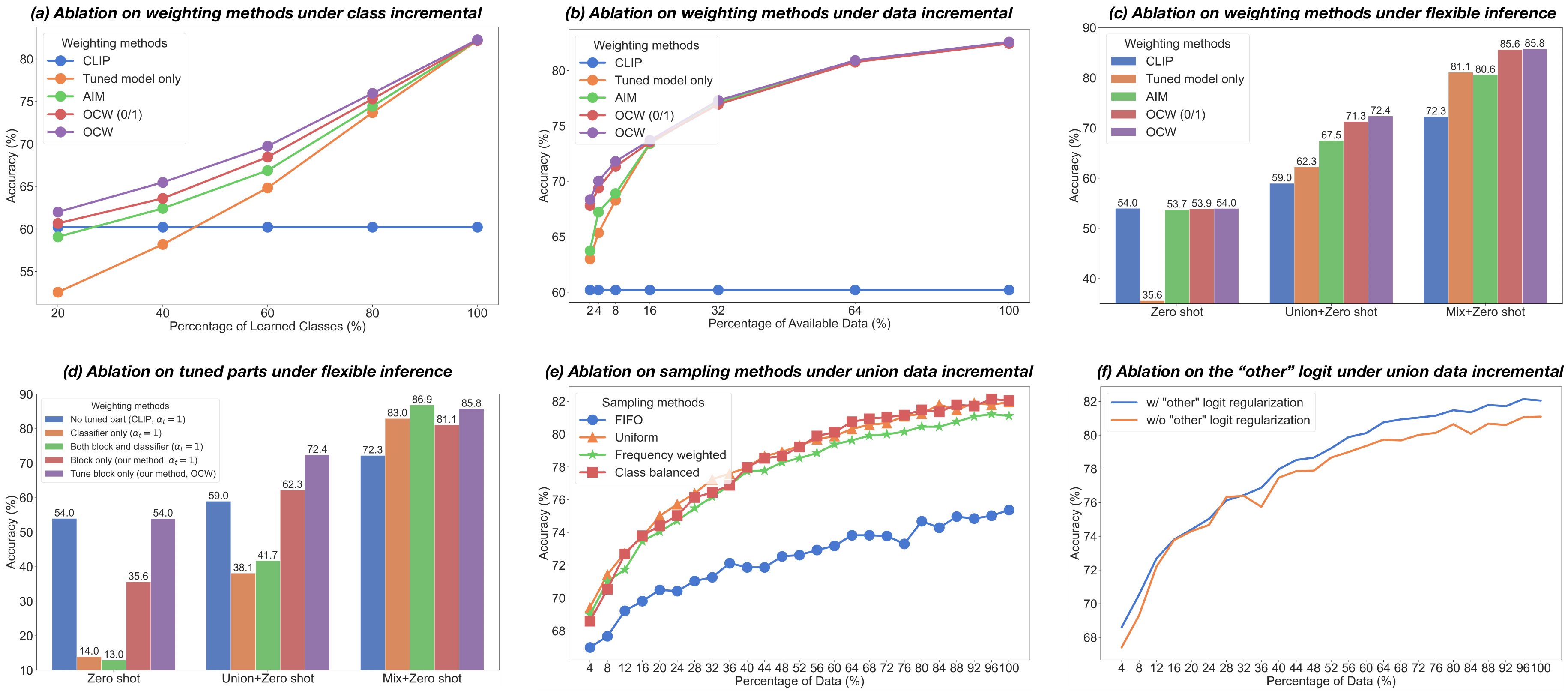}
\caption{Ablation results. Best viewed in color and zoomed-in.
\label{fig:ablation_experiments}
}
\end{minipage}
\end{figure*}
\label{sec:main_ablations}

{In the following, we ablate several important factors that contribute to our method. Additional ablations and results are provided in the supplemental. }

\noindent \textbf{Weighting Strategies}. 
Weighting is vital for our dual decoder approach. We compare several ways to compute the weights: 1) CLIP: namely $\alpha_t(y)=0$ for any images; 2) Tuned model only: $\alpha_t(y)=1$ for any images; 3) AIM~\cite{treeprobe}; 4) \weighting~(0/1): a variant of \weighting~where we round $\alpha_t(y)$ to 0 or 1 to use either the original or tuned model; 5) Our proposed \weighting. We partially finetune the decoder with fixed label embeddings and combine the tuned model with the original model using different weighting strategies. The results are shown in Fig.~\ref{fig:ablation_experiments} (a), (b) and (c), under data-, class-incremental and flexible inference. \weighting~performs the best of all methods in every setting and stage. Notably, \weighting~enables improvements even in the early stages when limited data is available for a class or task and better handles tasks that mix novel and target labels.



\noindent\textbf{Tuned parts}. Our proposed method tunes the last transformer block while keeping the label encoder fixed. In Fig.~\ref{fig:ablation_experiments} (d), we compare this approach to alternatives of tuning only the label encoder, both the block and the label encoder, or neither of them, under the flexible inference test.
When only using the tuned model for comparison ($\alpha_t=1$), fine tuning only the last transformer best retains predictive ability for novel labels. Further analysis under the task incremental scenario is provided in the supplemental material.



\noindent\textbf{Sampling methods.}~ 
We compare different methods for sampling from the stored samples. FIFO cycles through samples in order of first appearance, ``uniform'' randomly draws samples for each batch, class-balanced (which we use in all other experiments) samples classes, and frequency weighted sampling (FWS) samples based on how many times a given sample has been batched in training. Class-balanced and uniform are similar in practice, and perform best (Fig.~\ref{fig:ablation_experiments} (e)). This result is consistent with Prabhu~\etal.~\cite{prabhu_budgetedCL_cvpr_2023}'s finding that simple sampling methods outperform complex ones under a computational budget. See supplemental for more details.

\noindent \textbf{The ``other'' logit regularization}. The ablation experiments depicted in Fig.~\ref{fig:ablation_experiments} (f) assess the impact of ``other'' logit regularization in the union data incremental scenario. The results demonstrate consistent enhancements when this regularization is applied, compared to its absence.

\subsection{Scalability}
\label{sec:result_scalability}

\begin{figure*}[!htb]
 \centering
 \begin{minipage}{\textwidth}
\includegraphics[width=\linewidth]{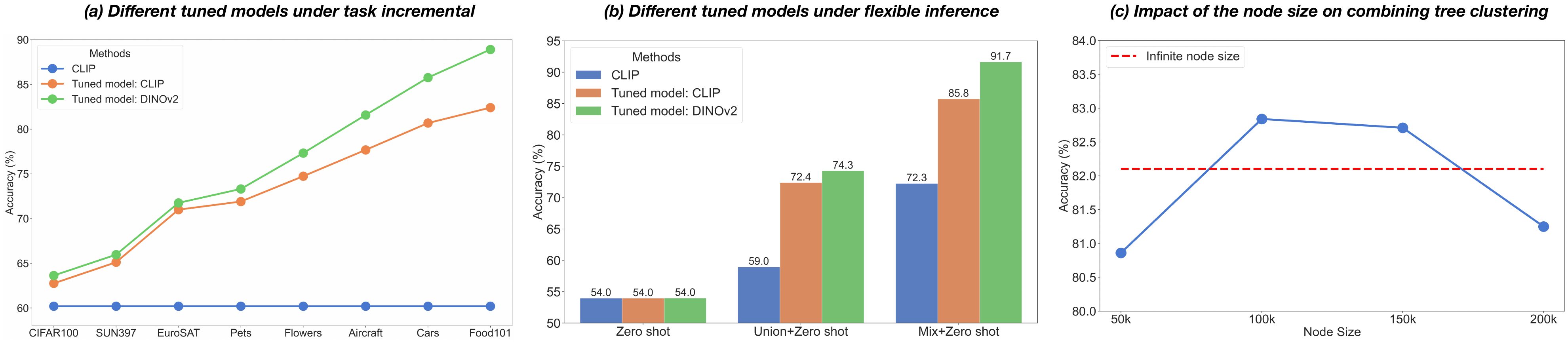}
\caption{(a) \& (b): The ``CLIP'' method refers to the original model. For Tuned model: DINOv2, we use the \href{https://github.com/facebookresearch/dinov2}{ViT-B/14 checkpoint}; (c): Infinite node size indicates only one tuned model regardless of the number of samples received. \label{fig:scalability_experiment}
}
\end{minipage}
\end{figure*}

\noindent \textbf{DINOv2 as the Tuned Model}. Using self-supervised DINOv2~\cite{Oquab_arxiv_2023_DINOv2} as our tuned model showcases our method's adaptability beyond the original CLIP model~\cite{treeprobe}. Our approach allows for tuning the feature embedding space while keeping label embeddings fixed, enabling the incorporation of diverse encoders through an additional linear layer for dimension matching. This layer is trained in joint with the tuned decoder. As illustrated in Fig.~\ref{fig:scalability_experiment} (a), replacing the tuned model with DINOv2 results in consistent performance improvements at every stage, with a notably steeper improvement curve in later stages. Furthermore, Fig.~\ref{fig:scalability_experiment} (b) demonstrates that this modification preserves Zero-shot performance while significantly boosting Union+Zero-shot and Mix+Zero-shot outcomes, attributed to DINOv2's enhanced performance on target tasks.


\noindent\textbf{Tree clustering models}. 
Zhu~\etal.~\cite{treeprobe} propose tree-based clustering as a way to limit the training time needed to incorporate new examples. Our method can already incorporate an example much faster in a way that does not depend on the number of training samples, but the clustering idea is still intriguing as a path to scalability in storage, memory, and model capacity. 
In Fig.~\ref{fig:scalability_experiment} (b), we evaluate different node capacities and find that moderate-sized partitions may slightly increase performance.  In these experiments, each node stores a copy of the last CLIP transformer block and updates it based on data that it receives.

\section{Discussion}

Our AnytimeCL approach shows promise of breaking free of staged train and deploy processes, with many benefits to applications that constantly or sporadically receive new data and have customizable or evolving label spaces. 

The {\em anytime} has three main parts: (1) being able to do a task even when training data do not fully cover the label space for that task; (2) being able to incorporate a new training example quickly; and (3) not losing the benefit of older training data.  Our key to the first part is using complimentary models. For that we are motivated by CLS theory~\cite{cls_oreilley_2014} and more specifically adopt the framework of Zhu \etal.~\cite{treeprobe}, improving it with our prediction weighting method and by partially fine-tuning the encoder.   Our key to the second part is that we do not need to incorporate one million examples quickly, just one.  This reframes the computationally budgeted CL problem of Prabhu \etal.~\cite{prabhu_budgetedCL_cvpr_2023}, who aim to incorporate a new large batch of examples in time that allows visiting only a portion of the existing data.  For some settings, our goal is more practical, as it enables an opportunistic approach that interleaves pure online learning during busy times and offline learning when time allows. While humans are often held up as excellent continual learners, they also learn one thing fast but many things slowly over time and make use of restful wake and sleep to continue learning. The effectiveness of our online learning is also due, in part, to leveraging a strong vision-language model, just recently possible. The key to the third goal is simple --- do not throw out the data.  If one cares about privacy and storage, then finding ways of compressing and preserving privacy, like our PCA-based approach, may be better than trying to retain the benefits of a large dataset without one.  

Potential directions for future work:   
\begin{itemize}
\item Beyond classification: Our partial fine-tuning approach is applicable to other tasks, such as semantic segmentation, visual question answering, and object detection. Our weighting method is applicable to open vocabulary detection and segmentation. 
\item Multi-model inference: Extending our approach to multiple models, including task-specific models, would further extend flexible learning and model re-use opportunities.
\item Scalability: We expect that tree-based data clustering and training one model per cluster, as in~\cite{treeprobe}, provides a good mechanism for scalability. Larger scale experiments, involving dozens of datasets and many millions of examples, are needed to fully test this.
\item Federated learning: Combining tree-based clustering and feature compression, we can encode training examples in the client or central server and transmit to a node that stores and updates the tunable block and its data. This enables fully distributed training on inexpensive nodes. \end{itemize}

\section{Conclusion}

We propose an effective approach for anytime continual learning of open vocabulary image classification.  The main innovation is dynamic class-senstive weighting for combining predictions of open-vocabulary and example-based tuned models, and we also find that partial fine-tuning with sufficiently small learning rates enables a surprisingly effective online learner.  We offer a per-image attention-weighted PCA approach to compress features, with benefits to storage, computation, and privacy. Our experiments show that our approach is very promising, and further exploration is merited to more fully understand its effective scope and limitations.


\section*{Acknowledgement}
This work is supported in part by ONR award N00014-21-1-2705, ONR award N00014-23-1-2383, and U.S.~DARPA ECOLE Program No.~\#HR00112390060. 
The views and conclusions contained herein are those of the authors and should not be interpreted as necessarily representing the official policies, either expressed or implied, of DARPA, ONR, or the U.S. Government. 

%
%
\bibliographystyle{splncs04}
\bibliography{main}

\clearpage
\setcounter{section}{0}
\setcounter{equation}{0}
\setcounter{figure}{0}
\setcounter{table}{0}
\makeatletter
\renewcommand{\thesection}{S-\arabic{section}}
\renewcommand{\theequation}{S\arabic{equation}}
\renewcommand{\thefigure}{S\arabic{figure}}
\renewcommand{\thetable}{S\arabic{table}}
\setcounter{page}{1}
\title{Anytime Continual Learning for Open Vocabulary Classification}
\author{Zhen Zhu\orcidlink{0000-0003-1557-8473} \and Yiming Gong\orcidlink{0009-0007-5520-7999} \and Derek Hoiem\orcidlink{0000-0001-6260-5708}}
\institute{University of Illinois at Urbana-Champaign\\
\email{\{zhenzhu4,yimingg8,dhoiem\}@illinois.edu}}
\maketitle

\section{Summary of contents}
The supplemental file contains:
\begin{itemize}
    \item {\bf Additional method details (Sec.~\ref{sec:sup_additional_method_details})}, including compression method details; the implementation details of the frequency weighted sampler.
    \item {\bf Additional ablation results (Sec.~\ref{sec:sup_ablation})}, regarding the online learning batch size selection; the EMA decay ($\eta$) for accuracy estimation in OCW; loss balancing parameter $\beta$ in Eq.~\ref{eq:final_loss}; the number of saved components for per-instance PCA compression; different per-instance PCA compression methods in the union data incremental scenario.
    \item {\bf Additional results to the main draft (Sec.~\ref{sec:additional_result_to_main_draft})}, including DINOv2~\cite{Oquab_arxiv_2023_DINOv2} vs. CLIP~\cite{radford_icml2021_clip} tuned model on class, data, and union data incremental; weighting method ablation on task incremental; tuned parts ablation on task incremental; detailed results of our method on the ZSCL~\cite{ZSCL} MTIL task incremental benchmark.
\end{itemize}

\section{Additional method details}
\label{sec:sup_additional_method_details}

\subsection{Compression method detail}
\label{sec:compression_detail}

In this section, we complement more details of methods and implementation of Tab.~\ref{tab:compression} in the main paper.
 
\noindent \textbf{Full Image.}~CLIP reshapes the input image to $224\times 224$ and then creates $32\times32$ non-overlapping image patches, resulting in $7\times 7$ patch tokens $\mathbf{P} \in \mathbb{R}^{49\times D}$ for the transformer, where $D$ (=768 in our case) represents the feature vector dimension. Additionally, a CLS token $\mathbf{C}\in \mathbb{R}^{1 \times D}$ is attached to the start of the token sequence. The CLS token after being processed through the whole transformer is normally followed by a classification layer in ViT~\cite{ViT} or directly used for calculating cosine similarity with label embeddings in CLIP~\cite{radford_icml2021_clip} (See Eq.~\ref{eq:class_wise_probability}). For fair comparison with other methods in Tab.~\ref{tab:compression}, we partially fine-tune the last layer of the transformer block without tuning the label embeddings. Processing a batch includes image loading, processing, a complete forward pass, and a backward pass restricted to the final transformer block.

\noindent \textbf{Full Features.}~We pre-compute the intermediate features $\mathbf{f}=[\mathbf{C}; \mathbf{P}] \in \mathbb{R}^{50\times D}$ before the final layer on all data and then store them to disk.  For fine-tuning, we load stored features back from disk to RAM and feed them directly into the final transformer block, which is the only block tuned. Processing a batch includes loading intermediate features, forward and backward pass of the final block.

\noindent \textbf{VQ.}~For Vector Quantization, we pre-train a codebook on intermediate features before the final layer using MSE loss between the quantized features and the original features. We adopt Zheng~\etal.~\cite{cvq-vae_iccv_2023}'s method to enhance VQ performance by promoting the use of more codebook vectors. Then, we process all images in the dataset to obtain the corresponding codebook indices. We store the codebook and indices of all samples to disk. Similar to Full Features, when fine-tuning, data processing involves loading a codebook and integer indices of intermediate features into RAM for feature reconstruction. To process a single batch, we need loading the codebook and indices, feature reconstruction by retrieving codewords from the codebook according to the indices, forward and backward pass of the final layer.

\noindent \textbf{PCA.}~To compress $\mathbf{f}$, we first center $\mathbf{f}$ by subtracting the mean values of each token: $\hat{\mathbf{f}} = \mathbf{f} - \mu$, where $\mu$ represent the mean values of all tokens. We follow scikit-learn to perform SVD: $\hat{\mathbf{f}} = \mathbf{U} \mathbf{\Sigma} \mathbf{V}^T$. The reduced form of $\hat{\mathbf{f}}$ can be approximated by $\hat{\mathbf{f}}\approx \mathbf{U}_n \mathbf{\Sigma}_n \mathbf{V}_n^T$, where $n$ is the number of principal components/singular values chosen. $\mathbf{U}_n \mathbf{\Sigma}_n $ is the PCA coefficient matrix and $\mathbf{V}_n^T$ is the PCA component matrix. Ideally, we want to find a small $n$ so that the accuracy after fine-tuning is reasonably good. We store $\mathbf{U}_n \mathbf{\Sigma}_n$, $\mathbf{V}_n$ and $\mu$ to the disk for reconstructing $\mathbf{f}$ during training: $\mathbf{U}_n \mathbf{\Sigma}_n \mathbf{V}_n^T + \mu$.

\noindent \textbf{Dataset-wide PCA.}~Since $\mathbf{f}$ is of shape $50\times D$ and in our case $D=768>50$, it makes a higher compression rate to compress along the vector dimension. For dataset-wide PCA, we concatenate the intermediate features from all samples together along the dimension of the 50 tokens and then apply PCA on the whole concatenated feature. Due to hardware limitation of RAM capacity, processing all data in a single pass is not feasible. Therefore, we divide all data into chunks of 5,000 samples each. PCA is performed on each chunk and we store feature means, PCA coefficients, and components to disk. Processing a single batch involves loading these data from disk to RAM, reconstructing features via PCA, and processing them through the final layer (both forward and backward pass).

\noindent \textbf{Per-instance PCA.}~For this method, PCA is applied on the 50 tokens of {\em each sample}. We store the feature means, PCA coefficients, and components to disk. Procedures required to process a batch are similar as dataset-wide PCA.

\noindent \textbf{CLS-weight.}~We use the CLS token $\mathbf{C}$ for classification purposes. The idea is that some of the patch tokens $\mathbf{P}$ are quite similar to this CLS token and could be very important for the classification task. To take advantage of this, we adjust the importance (or weight) of these patch tokens based on how similar they are to the CLS token. By doing so, we can potentially compress the information more effectively without losing important details necessary for accurate classification. Specifically, these similarities are obtained through:
\begin{equation}
    S = \text{Softmax}(\text{Matmul}(\text{LayerNorm}(\mathbf{C}), \text{LayerNorm}(\mathbf{P}^{{T}}))).
\end{equation}
Here, we use the layer normalization from the last transformer block. 
Then, the patch tokens are re-weighted by: $S \cdot \mathbf{P}$.
Following this reweighting, per-instance PCA is applied to the adjusted intermediate features: $f_{\text{weighted}}=[\mathbf{C};S \cdot \mathbf{P}]$.

\noindent \textbf{Int-quantization.}~To further reduce memory usage, we convert principal components from 32-bit floats to 8-bit integers, potentially reducing memory by approximately fourfold. Suppose a principal component is $V$, we map it to uint8 (0–255) after performing min-max normalization:
\begin{equation}
    \hat{V} = \text{Int}(255 * (\frac{V - V_{\text{min}}}{V_{\text{max}} - V_{\text{min}}})),
\end{equation}
where $V_{\text{min}}$ and $V_{\text{max}}$ are the minimum and maximum values of vector $V$; $\text{Int}(\cdot)$ means converting the values of a vector to integers.
After the quantization process, we store the min, max values, and the uint8 vector. Reconstruction involves mapping the vector back to float numbers. A similar process can be applied to principal coefficients, although the decrease in storage is less significant.

\noindent \textbf{Implementation details for obtaining the numbers in Tab.~\ref{tab:compression}.}~We run the same test 100 times for each method without other ongoing programs, and then compute the mean as the result of the time needed to process a single batch. We perform partial fine-tuning on the final block of the transformer network. We opt for a batch size of 32, a learning rate of 5e-6, a weight decay of 0.05, and conduct fine-tuning over 10 epochs.

\subsection{Frequency-weighted sampling method details}

The idea of {\bf Frequency Weighted Sampling} (FWS) is to assign more sampling weights for recent samples while ensuring that older samples can be selected. Specifically, when a new sample $x_i$ is received, its sampling weight $w_i$ is initialized to 1. After including the sample in a training batch, the weight is updated by:
\begin{equation}
    w_i=\max(w_i * \xi, w_{\min}),
\end{equation}
where $\xi \in [0, 1]$ is the decay multiplier.  To be clear, $w_i$ is proportional to the likelihood that a training sample is used in the next training batch, not as a loss weight.  We set a lower bound $w_{\min}$ for sampling weights, for numerical stability and to ensure relatively well-trained instances have equal chance of being sampled.  When a new training example is received, a batch of size $B$ is filled with that example and $B-1$ other random stored examples, drawn with a $w_i$ weighting. One training iteration is performed with that batch. The FWS strategy hastens convergence to the expectation that all training samples are eventually sampled an equal number of times, regardless of their order of appearance. $\xi$ is not critical to performance: whether it is 0 or 1, it is equivalent to uniform sampling; any values in between adjust the degree of frequency bias---larger values lead to more repeated sampling of recently added examples. In the comparison, we set it to 0.99.

\subsection{More implementation details}
If not otherwise specified, we use the same seed for all experiments for consistency. Our dataset splits on class, data, task incremental, and flexible inference are the same as Zhu~\etal.~\cite{treeprobe}, for fair comparison. 
The offline AnytimeCL model is trained with 10 epochs for each stage. 

\section{Additional ablation experiments}
\label{sec:sup_ablation}

\begin{figure*}[!htb]
 \centering
 \begin{minipage}{\textwidth}
\includegraphics[width=\linewidth]{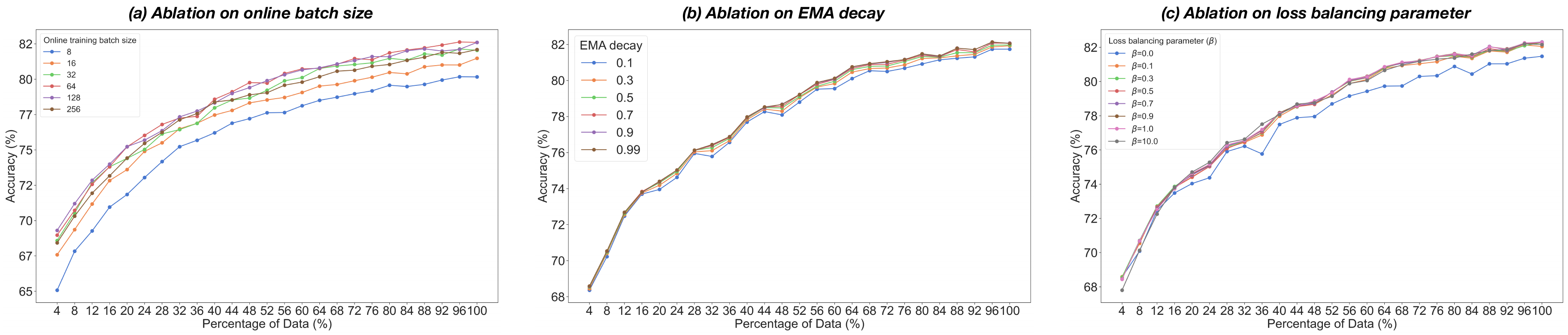}
\caption{Additional ablation results. Best viewed in color and zoomed-in. All experiments in these plots are conducted under the union data incremental setting. Our default selected options are 32 as the online training batch size in (a), 0.99 as the EMA decay for accuracy estimation in (b), 0.1 as the loss balancing parameter in (c). 
\label{fig:supplemental_ablation}
}
\end{minipage}
\end{figure*}

\subsection{Ablation on the online training batch size}
For our online training method, assume the batch size is $B$, the total computation allowed for training new samples equals to processing $B$ epochs of new samples.
Therefore, our online learning method can fit to different computation budgets by using different online learning batch sizes. We evaluate the impact of batch size on union data incremental in Fig.~\ref{fig:supplemental_ablation} (a). For this comparison, we keep the ratio of learning rate and batch size constant. When the batch size is smaller than 32, the corresponding accuracy, especially in later portions of data received, is obviously lower than other batch sizes. When the batch size is 64 or 128, the performance can be better than using 32. However, the performance of batch size 256 is relatively worse than 32. This shows that given larger computation budget (\ie, increasing batch size), performance can be improved though the gain diminishes quickly. 

\subsection{Ablation on the EMA decay for the accuracy estimation of OCW}

As can be seen in Fig.~\ref{fig:supplemental_ablation} (b), our method is robust against different EMA decays but generally larger EAM decay leads to better result. As a result, we choose 0.99 as its default value throughout all experiments.

\subsection{Ablation on the loss balancing parameter $\beta$}

In Sec.~\ref{sec:main_ablations} of the main paper, we already showed using the ``other'' logit regularization enhances the performance. Here in Fig.~\ref{fig:supplemental_ablation} (c), we additionally test the performance of various loss balancing weights. The results validate including the regularization term is important but the exact value of the loss balancing parameter is not critical to performance.



\subsection{Ablation on the number of components for per-instance PCA}

As default, we use only 5 components for per-instance PCA in Tab.~\ref{tab:compression}. Here, we additionally present results using 3, 10, and 20 components in Tab.~\ref{tab:different_components}. Using 5 components maintains good accuracy after applying our proposed techniques (+CLS-weight and int-quantize). It is possible to compress further by using 3 components but the average accuracy also decreases. The gains of introducing more components than 10 are not significant. 



\begin{table*}
\centering
\scalebox{0.89}{
\begin{tabular}{l|cc|cc|cc}
\toprule
No. components & \multicolumn{2}{c|}{Per-instance} & \multicolumn{2}{c|}{+CLS-weight} & \multicolumn{2}{c}{+int-quantize} \\
 & FT Acc. & KB/sample & FT Acc. & KB/sample & FT Acc. & KB/sample \\
\midrule
3 & $75.2 \pm 0.23$ & 12.9 & $77.4 \pm 0.26$ & 12.9 & $77.4 \pm 0.22$ & 3.7 \\
5 & $74.7 \pm 0.34$ & 19.4 & $77.7 \pm 0.19$ & 19.4 & $77.6 \pm 0.16$ & 5.3 \\
10 & $77.4 \pm 0.17$ & 35.8 & $78.1 \pm 0.14$ & 35.8 & $77.7 \pm 0.25$ & 9.4 \\
20 & $77.7 \pm 0.11$ & 68.5 & $77.8 \pm 0.35$ & 68.5 & $77.7 \pm 0.20$ & 17.7 \\
\bottomrule
\end{tabular}
}
\caption{Fine-tuning accuracy for CIFAR100 when choosing different number of components. FT accuracy is obtained from averaging six runs with different seeds. Numbers after $\pm$ denote the standard deviations of FT accuracies across six runs.}
\label{tab:different_components}
\end{table*}


\subsection{Ablation on the per-instance PCA compression methods under union data incremental}
\begin{figure*}[!htb]
 \centering
 \begin{minipage}{\textwidth}
\includegraphics[width=\linewidth]{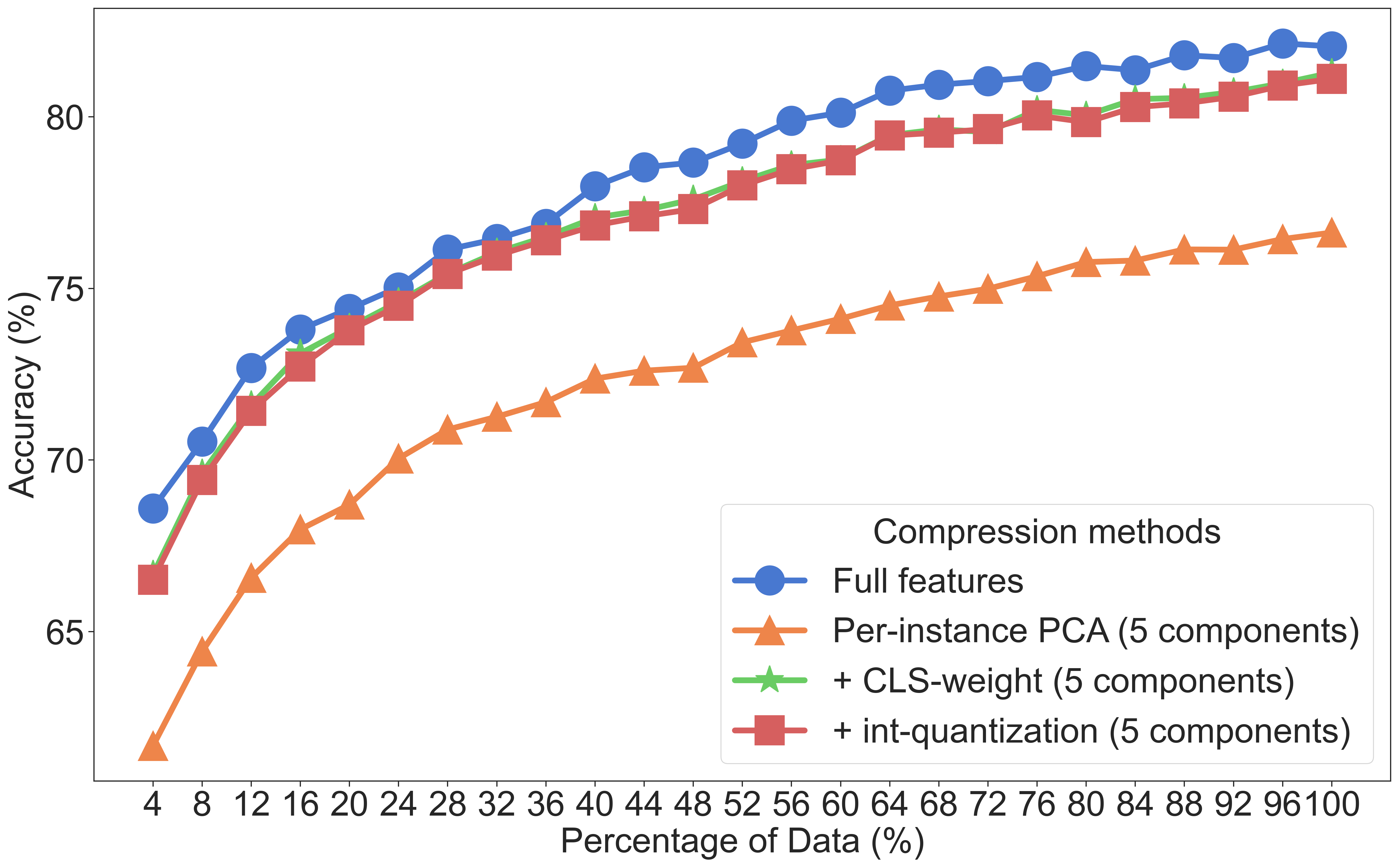}
\caption{Comparison of different per-instance PCA compression methods under the union data incremental scenario.
\label{fig:per_instance_pca_union_data_incremental}
}
\end{minipage}
\end{figure*}
To conduct the comparison to using full features on the union data incremental scenario, for the per-PCA compression methods, we load all cached contents to RAM and recover the intermediate features on the fly. Using full features requires extensive I/O operations to load stored features from disk during training, though it no reconstruction is needed.
As presented in Fig.~\ref{fig:per_instance_pca_union_data_incremental}, from comparing per-instance PCA and +CLS-weight, reweighting the patch tokens using their similarities to the CLS token dramatically improves performance. The int quantization technique compresses the data further without compromising on performance. Both +CLS-weight and +int-quantization are close to the full features method in every stage, showing the effectiveness of our proposed compression methods.

\section{Additional results to the main draft}
\label{sec:additional_result_to_main_draft}

\subsection{DINOv2~\cite{Oquab_arxiv_2023_DINOv2} vs. CLIP~\cite{radford_icml2021_clip} tuned model on class, data, and union data incremental}

\begin{figure*}[!htb]
 \centering
 \begin{minipage}{\textwidth}
\includegraphics[width=\linewidth]{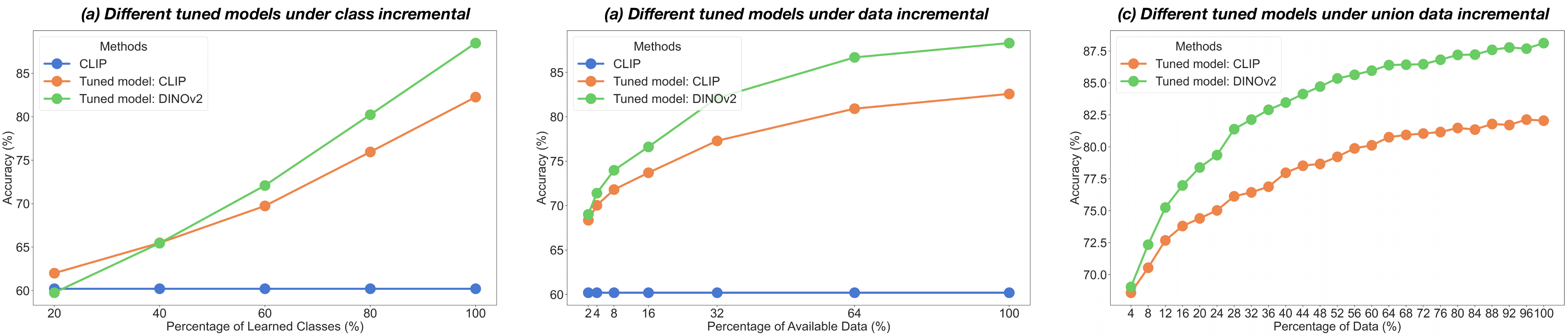}
\caption{Comparison between DINOv2~\cite{Oquab_arxiv_2023_DINOv2} and CLIP~\cite{radford_icml2021_clip} as tuned model on class (a), data (b), and union data incremental scenarios (c). 
\label{fig:supplemental_dino_scalability_results}
}
\end{minipage}
\end{figure*}

Fig.~\ref{fig:supplemental_dino_scalability_results} shows the comparison. Clearly using DINOv2 as the tuned model has better results eventually in all cases. It is interesting that DINOv2 as the tuned model shows superiority over CLIP in the very beginning of data incremental scenarios in Fig.~\ref{fig:supplemental_dino_scalability_results} (b) \& (c), unlike the class incremental case in (a). This result may indicate that DINOv2 is more helpful in few-shot learning scenarios.

\begin{figure*}[!htb]
 \centering
 \begin{minipage}{\textwidth}
\includegraphics[width=\linewidth]{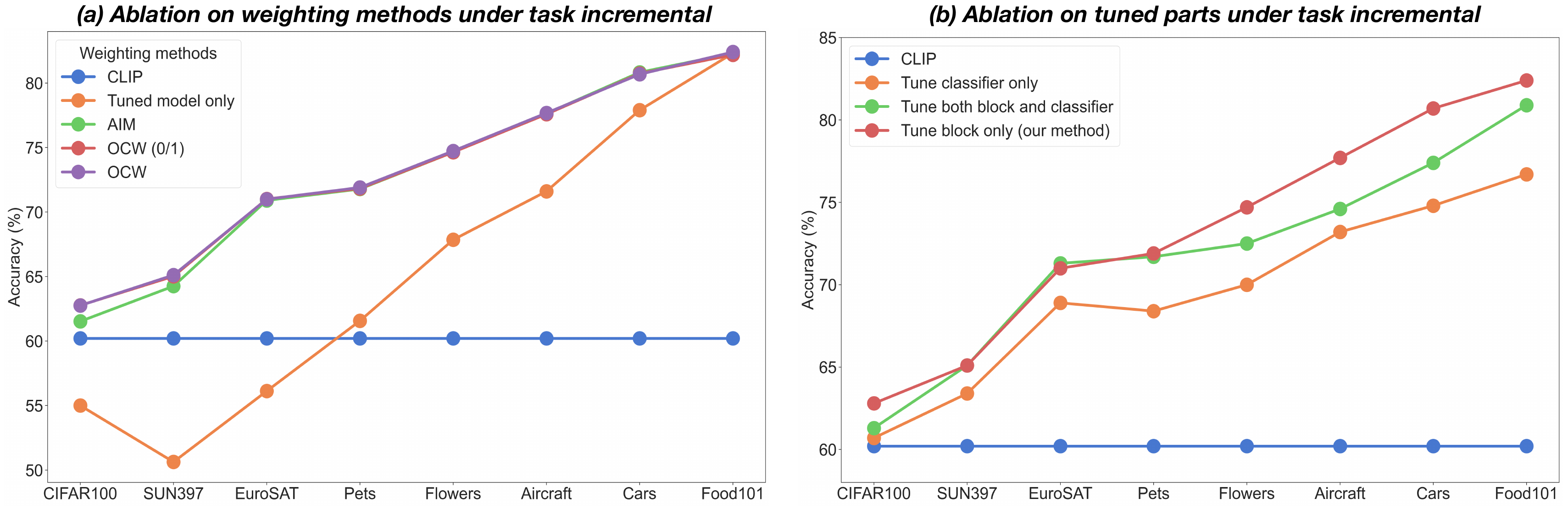}
\caption{Supplemental results to existing ablations in the main draft. In (a), OCW (0/1) is hard to be observed since OCW overlaps with it. 
\label{fig:supplemental_results_to_existing_ablations}
}
\end{minipage}
\end{figure*}

\subsection{Weighting method ablation on task incremental}
Fig.~\ref{fig:supplemental_results_to_existing_ablations} (a) shows the results of different weighting methods on the task incremental scenario. This scenario cannot fully demonstrate the effectiveness of weighting methods for the similar reason stated in Sec.~\ref{sec:zscl_comparison}.

\subsection{Tuned parts ablation on task incremental}
Our proposed method only tunes the last transformer block and freezes the label encoder. This leads to better target task performance than replacing the label encoder with a tuned classifier, as shown in Fig.~\ref{fig:supplemental_results_to_existing_ablations} (b). Training only the classifier layer underperforms because it cannot tune the features. When fine tuning the last transformer block, keeping the label encoder fixed improves performance, perhaps because it provides a regularization on the feature tuning.

\subsection{Detailed results under the MTIL task incremental learning benchmark~\cite{ZSCL}}

We also present the detailed accuracies of AnytimeCL online and offline in Tab.~\ref{tab:anytimeCL_online_zscl} and Tab.~\ref{tab:anytimeCL_offline_zscl}, respectively. Our internal tests revealed that hyperparameter tuning, such as adjusting the learning rate for each task, can enhance results, as revealed in {\href{https://github.com/Thunderbeee/ZSCL/tree/8cd0caf19dfbb024dcde3fb3af8adbf6949259e8}{ZSCL's released code}}~\cite{ZSCL}. However, we tend to avoid hyperparameter tuning of different tasks since selected hyperparameters do not guarantee a better performance when the data distribution in an online data stream shifts.

\begin{table*}
\centering
\resizebox{\textwidth}{!}{
    \begin{tabular}{@{}lcccccccccccc@{}}
    \toprule
    \noalign{\smallskip}
     & \rotatebox{90}{Aircraft} & \rotatebox{90}{Caltech101} & \rotatebox{90}{CIFAR100} & \rotatebox{90}{DTD} & \rotatebox{90}{EuroSAT} & \rotatebox{90}{Flowers} & \rotatebox{90}{Food} & \rotatebox{90}{MNIST} & \rotatebox{90}{OxfordPet} & \rotatebox{90}{Cars} & \rotatebox{90}{SUN397} & \\ 
    \noalign{\smallskip} 
    \hline \hline \noalign{\smallskip}
    Transfer &  & 87.9 & 68.2 & 45.3 & 54.6 & 71.4 & 88.9 & 59.4 & 89.1 & 64.6 & 64.1 & \cellcolor{green!40}{69.4} \\
    \midrule
    Aircraft & 44.85 & 87.90 & 68.22 & 45.32 & 54.61 & 71.43 & 88.86 & 59.45 & 89.07 & 64.61 & 64.05\\
    Caltech101 & 50.50 & 96.60 & 68.22 & 45.32 & 54.61 & 71.43 & 88.86 & 59.45 & 89.07 & 64.61 & 64.05\\
    CIFAR100 & 52.45 & 96.89 & 82.23 & 45.32 & 54.61 & 71.43 & 88.86 & 59.45 & 89.07 & 64.61 & 64.05\\
    DTD & 52.42 & 96.66 & 83.03 & 69.63 & 54.61 & 71.43 & 88.86 & 59.45 & 89.07 & 64.61 & 64.05\\
    EuroSAT & 52.78 & 96.77 & 83.57 & 75.64 & 94.46 & 71.43 & 88.86 & 59.45 & 89.07 & 64.61 & 64.05 \\
    Flowers & 53.59 & 96.83 & 83.52 & 74.95 & 95.59 & 87.84 & 88.86 & 59.45 & 89.07 & 64.61 & 64.05\\
    Food & 54.04 & 96.77 & 83.6 & 75.11 & 96.63 & 92.83 & 91.36 & 59.45 & 89.07 & 64.61 & 64.05\\
    MNIST & 54.40 & 96.49 & 83.77 & 75.32 & 96.19 & 93.23 & 91.60 & 98.51 & 89.07 & 64.61 & 64.05 \\
    OxfordPet & 55.12 & 96.43 & 83.54 & 75.37 & 96.83 & 92.97 & 92.22 & 98.76 & 91.63 & 64.61 & 64.05\\
    Cars & 53.44 & 96.60 & 83.68 & 74.73 & 96.63 & 92.94 & 92.10 & 98.58 & 92.75 & 83.48 & 64.05\\
    SUN397 & 53.11 & 96.37 & 83.27 & 73.51 & 95.93 & 92.88 & 92.04 & 98.36 & 93.16 & 85.77 & 79.67 & \cellcolor{orange!40}{85.8} \\
    \midrule
    Avg. & 52.4 & 95.8 & 80.6 & 66.4 & 81.0 & 82.7 & 90.2 & 73.7 & 90.0 & 68.2 & 65.5 & \cellcolor{blue!30}{77.0}\\
    \bottomrule
    \end{tabular}
}
\caption{Accuracy (\%) of AnytimeCL Online on the MTIL benchmark with order-I. Each row represents the performance on every dataset of the model trained after the corresponding task. Transfer, \cellcolor{blue!30}{Avg.}, and Last metrics are shown in color. We follow the same table arrangement as in ZSCL~\cite{ZSCL}.}
\label{tab:anytimeCL_online_zscl}
\end{table*}

\begin{table*}
\centering
\resizebox{\textwidth}{!}{
    \begin{tabular}{@{}lcccccccccccc@{}}
    \toprule
    \noalign{\smallskip}
     & \rotatebox{90}{Aircraft} & \rotatebox{90}{Caltech101} & \rotatebox{90}{CIFAR100} & \rotatebox{90}{DTD} & \rotatebox{90}{EuroSAT} & \rotatebox{90}{Flowers} & \rotatebox{90}{Food} & \rotatebox{90}{MNIST} & \rotatebox{90}{OxfordPet} & \rotatebox{90}{Cars} & \rotatebox{90}{SUN397} & \\ 
    \noalign{\smallskip} 
    \hline \hline \noalign{\smallskip}
    Transfer &  & 87.9 & 68.2 & 45.3 & 54.6 & 71.4 & 88.9 & 59.4 & 89.1 & 64.6 & 64.1 & \cellcolor{green!40}{69.4} \\
    \midrule
    Aircraft & 36.87 & 87.90 & 68.22 & 45.32 & 54.61 & 71.43 & 88.86 & 59.45 & 89.07 & 64.61 & 64.05\\
    Caltech101 & 45.48 & 97.00 & 68.22 & 45.32 & 54.61 & 71.43 & 88.86 & 59.45 & 89.07 & 64.61 & 64.05\\
    CIFAR100 & 49.68 & 96.72 & 84.46 & 45.32 & 54.61 & 71.43 & 88.86 & 59.45 & 89.07 & 64.61 & 64.05\\
    DTD & 51.28 & 96.77 & 84.14 & 73.83 & 54.61 & 71.43 & 88.86 & 59.45 & 89.07 & 64.61 & 64.05\\
    EuroSAT & 52.18 & 96.71 & 84.48 & 74.79 & 97.37 & 71.43 & 88.86 & 59.45 & 89.07 & 64.61 & 64.05 \\
    Flowers & 53.02 & 96.89 & 84.45 & 75.16 & 97.59 & 91.67 & 88.86 & 59.45 & 89.07 & 64.61 & 64.05\\
    Food & 52.39 & 96.95 & 84.20 & 74.41 & 97.54 & 91.72 & 92.43 & 59.45 & 89.07 & 64.61 & 64.05\\
    MNIST & 53.17 & 96.66 & 84.3 & 74.73 & 97.59 & 91.90 & 92.48 & 98.99 & 89.07 & 64.61 & 64.05 \\
    OxfordPet & 53.92 & 96.83 & 84.45 & 74.31 & 97.52 & 91.88 & 92.35 & 99.16 & 93.92 & 64.61 & 64.05\\
    Cars & 53.80 & 97.06 & 84.23 & 75.05 & 97.50 & 91.54 & 92.41 & 99.20 & 94.11 & 85.86 & 64.05\\
    SUN397 & 52.48 & 96.72 & 83.83 & 73.94 & 97.33 & 91.56 & 92.23 & 99.14 & 93.73 & 85.84 & 81.50 & \cellcolor{orange!40}{86.2} \\
    \midrule
    Avg. & 50.4 & 96.0 & 81.3 & 66.6 & 81.9 & 82.5 & 90.5 & 73.9 & 90.4 & 68.5 & 65.6 & \cellcolor{blue!30}{77.0}\\
    \bottomrule
    \end{tabular}
}
\caption{Accuracy (\%) of AnytimeCL Offline on the MTIL benchmark with order-I. 
}
\label{tab:anytimeCL_offline_zscl}
\end{table*}

\end{document}